\documentclass[10pt,a4paper]{article}

\usepackage{caption}
\usepackage{subcaption}
\usepackage{graphicx}
\graphicspath{{figures/}}

\usepackage{amssymb}
\usepackage{amsmath} 

\usepackage{array}
\usepackage{lipsum}
\usepackage{hyperref}
\usepackage{authblk}
\usepackage{setspace}
\usepackage{algorithm}
\usepackage{algpseudocode}
\usepackage{multirow}
\usepackage{booktabs}

\begin{document}
	
	\title{Benchmark Functions for CEC 2022 Competition on
		Seeking Multiple Optima in Dynamic Environments } 
	\author[1]{Wenjian Luo}
	\author[1, 2]{Xin Lin}
	\author[3]{Changhe Li}
	\author[4]{Shengxiang Yang}
	\author[5]{Yuhui Shi}
	
	\affil[1]{School of Computer Science and Technology, Harbin Institute of Technology, Shenzhen 518055, China}
	\affil[2]{School of Computer Science and Technology, University of Science and Technology of China, Hefei 230027, Anhui, China}
	\affil[3]{School of Automation, China University of Geosciences, 388 Lumo Road,Wuhan, 430074, China}
	\affil[4]{School of Computer Science and Informatics, the De Montfort University, Leicester LE1 9BH, United Kingdom}
	\affil[5]{Department of Computer Science and Engineering, Southern University of Science and Technology, Shenzhen 518055, China}
	\affil[ ]{Email:
		\href{mailto:luowenjian@hit.edu.cn}{luowenjian@hit.edu.cn}, \href{mailto:iskcal@mail.ustc.edu.cn}{iskcal@mail.ustc.edu.cn}, 
		\href{mailto:change.lw@gmail.com}{change.lw@gmail.com},
		\href{mailto:syang@dmu.ac.uk}{syang@dmu.ac.uk},
		\href{mailto:x.yao@cs.bham.ac.uk}{shiyh@sustech.edu.cn}}
	\date{\today} 
	\maketitle 

	\begin{abstract}
		\normalsize
		Dynamic and multimodal features are two important properties and widely existed in many real-world optimization problems. The former illustrates that the objectives and/or constraints of the problems change over time, while the latter means there is more than one optimal solution (sometimes including the accepted local solutions) in each environment. The dynamic multimodal optimization problems (DMMOPs) have both of these characteristics, which have been studied in the field of evolutionary computation and swarm intelligence for years, and attract more and more attention. Solving such problems requires optimization algorithms to simultaneously track multiple optima in the changing environments. So that the decision makers can pick out one optimal solution in each environment according to their experiences and preferences, or quickly turn to other solutions when the current one cannot work well. This is very helpful for the decision makers, especially when facing changing environments. In this competition, a test suit about DMMOPs is given, which models the real-world applications. Specifically, this test suit adopts 8 multimodal functions and 8 change modes to construct 24 typical dynamic multimodal optimization problems. Meanwhile, the metric is also given to measure the algorithm performance, which considers the average number of optimal solutions found in all environments. This competition will be very helpful to promote the development of dynamic multimodal optimization algorithms.
		
	\end{abstract}
	
	
	\clearpage
	
	\setcounter{tocdepth}{2}
	\tableofcontents

	\clearpage

	\section{Introduction}
	\label{sec:intro}
	
	In the real-world applications, a large amount of optimization problems have the dynamic property, which is called dynamic optimization problems (DOPs) \cite{mavrovouniotis2017survey}, such as dynamic economic dispatch problems \cite{zou2022differential}, and dynamic load balancing problems \cite{sesum2010comparing}. The objectives and/or constraints of these problems vary over time. In the field of evolutionary computation and swarm intelligence, there exist lots of work dedicated to improving the adaption of evolutionary algorithms to the dynamic environments \cite{zhu2018global, cao2018neighbor, zhu2020making, luo2018surrogate, rong2018multidirectional}.
	
	
	The problems with multimodal nature, which is called multimodal optimization problems (MMOPs) \cite{das2011real} have also been studied for many years in the field of evolutionary computation and swarm intelligence. These problems have more than one global optimum or several accepted local optimal solutions. Basic stochastic optimization algorithms have no ability to catch all the optima of MMOPs. Instead, the niching method and multiobjective optimization method are widely adopted to solve MMOPs \cite{lin2019differential, wang2017dual, wang2014mommop, luo2020hybridizing}.
	
	Recently, the dynamic multimodal optimization problems (DMMOPs), which have both dynamic and multimodal features, have been attracting the researchers' attentions. Compared with DOPs, DMMOPs have more than one optima (sometimes including the accepted local optima) in every environment. Formula (\ref{eqt:dmmop}) shows the definition of DMMOPs,
	
	\begin{equation}
		\label{eqt:dmmop}
		\left(
		\begin{array}{c}
			\boldsymbol{o}_{11} \\
			\boldsymbol{o}_{12} \\
			\cdots \\
			\boldsymbol{o}_{1n_{1}} \\
		\end{array},
		\begin{array}{c}
			\boldsymbol{o}_{21} \\
			\boldsymbol{o}_{22} \\
			\cdots \\
			\boldsymbol{o}_{2n_{2}} \\
		\end{array},
		\cdots,
		\begin{array}{c}
			\boldsymbol{o}_{t1} \\
			\boldsymbol{o}_{t2} \\
			\cdots \\
			\boldsymbol{o}_{tn_{t}} \\
		\end{array}
		\right) = \mathop{\arg\max}_{\boldsymbol{x}\in\mathit{\Omega}}~f(\boldsymbol{x}, t),
	\end{equation}
	where $\boldsymbol{o}_{tn_{t}}$ represents the $n_t$-th optimal solution in the $t$-th environment. Problem $f(\boldsymbol{x}, t)$ is a maximized DMMOP. It can be seen that the optimization algorithms solving DMMOPs should find all the optimal solutions in each environment.
	
	Some algorithms have been proposed to deal with DMMOPs in the field of evolutionary computation and swarm intelligence. In \cite{luo2019clonal}, Luo \textit{et al.} designed an improved  clonal selection algorithm with a niching method and a memory strategy to solve DMMOPs. In \cite{cheng2018dynamic}, Cheng \textit{et al.} proposed a novel brain storm optimization with the new individual generation strategy to deal with the dynamic multimodal environment. In \cite{ahrari2021adaptive}, Ahrari \textit{et al.} reformed the covariance matrix self-adaption evolution strategy by an adaptive multilevel prediction method to find multiple optima in the dynamic environment. In \cite{cuevas2021evolutionary}, Cuevas \textit{et al.} proposed the mean shift scheme method to track optima in DMMOPs with the consideration of both density and fitness values of individuals. 
	

	
	A benchmark test problem designed from different perspectives is needed to check the performance of the algorithms. Recently, Lin \textit{et al.} \cite{lin2021popdmmo} proposed a general framework for solving DMMOPs, called PopDMMO. It considers multiple behaviors simulating various situations of DMMOPs. Specifically, 8 basic multimodal environments and 8 dynamic change modes are included in PopDMMO. From the perspective of multimodal property, PopDMMO adopts two categories of functions to simulate the fitness landscape with multiple optimal solutions. The first category is based on the DF functions \cite{morrison1999test}, which are the popular benchmarks in evolutionary dynamic optimization. Here, DF functions are adapted as the simple multimodal environment with several global and local peaks. The second is the composition functions in CEC 2013 competition on niching methods for multimodal function optimization \cite{li2013benchmark}. The landscape of the composition functions has a huge amount of local peaks which may mislead the evolutionary process of the optimization algorithms. From the perspective of dynamic property, PopDMMO contains 8 change behaviors to simulate the behaviors of dynamic changes. In detail, 6 change modes are derived from the generalized dynamic benchmark generator (GDBG) \cite{li2008benchmark}. These change behaviors are made by modifying key parameters for the changing of the fitness landscapes. Besides, 2 additional change behaviors are proposed in PopDMMO. These two changes mainly focus on the number of optimal solutions, i.e., linear change and random change.
	
	
	This report gives a set of complete definitions of DMMOPs, which is extracted from PopDMMO. The benchmark problems mainly focus on the various situations of the multimodal fitness landscape and different change modes of the dynamic nature. It is noted that all functions are maximized. The website for the competition is available at the following link.
	
	\centerline{ \url{http://mi.hitsz.edu.cn/activities/2022dmmo_competition.html}}
	
	
	In the following contents, Section \ref{sec:multi} describes the definitions of the basic multimodal landscapes. Section \ref{sec:dynamic} illustrates the change behaviors when the environmental change occurs. The experimental criteria are given in Section \ref{sec:ex}. 
	
	\section{Multimodal Functions}
	\label{sec:multi}
	
	This section mainly focuses on the multimodal fitness landscape, simulating different multimodal occasions. Eight multimodal functions are adopted in our report, and these functions are separated into two categories: the simple multimodal functions modeled by the DF generator \cite{morrison1999test}, and the complex multimodal functions constituted by the composition functions \cite{li2013benchmark}. Each category has 4 functions.
	
	\subsection{Simple Multimodal Functions}
	The simple multimodal functions are all based on the DF generator \cite{morrison1999test}. The fitness calculation is shown as follows.
	\begin{equation}
		\label{eqt:df1}
		\begin{split}
			\begin{aligned}
				f&(\boldsymbol{x}) = 
				& \max \limits_{\substack{i = 1 \cdots G, \\ G+1, \cdots G+L}} \left(H_i - W_i * \sqrt{\sum_{j=1}^{D}(x_j - X_{ij})^2}\right),
			\end{aligned}
		\end{split}
	\end{equation}
	where $H_i$ and $W_i$ represent the initial height and width of the $i$-th peak. Parameter $X_{ij}$ represents the initial $j$-th dimensional value of the $i$-th peak. Parameter $G$ and $L$ represent the number of global peaks and the maximum number of local peaks, respectively. It is noted that the heights of the global peaks are the same and higher than the local peaks, and the minimum distance between peaks should not be less than 0.1. Other parameters are randomly pre-defined.
	
	\subsubsection{DF Function $F_1$} 
	
	\begin{itemize}
		\item Property: all data are randomly initialized.
		
		\item Number of global optima: 4.
		
		\item Number of local optima: at most 4.
		
		\item Widths of peaks: randomly set.
		
		\item Heights of peaks: 75 of global peaks, randomly values smaller than 75 of local peaks.
		
		\item Positions of peaks: randomly set.
		
		\item Range for each dimension: [-5, 5].
	\end{itemize}
	
	\begin{figure}[h]
		\centering
		\includegraphics[width=0.65\columnwidth]{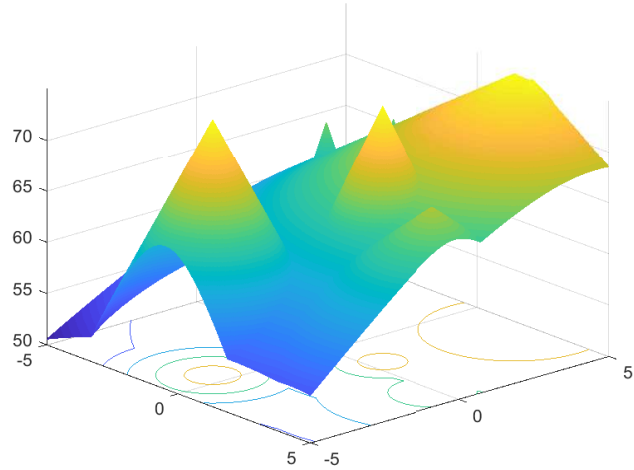}
		\caption{Fitness landscape of F1 in the 2-dimensional problem}	
	\end{figure}

	\subsubsection{DF Function $F_2$}
	
	\begin{itemize}
		\item Property: simulating F6 in CEC 2013 competition on niching methods for multimodal function optimization \cite{li2013benchmark}.
		
		\item Number of global optima: 4.
		
		\item Number of local optima: 0.
		
		\item Widths of peaks: 12.
		
		\item Heights of peaks: 75.
		
		\item Positions of peaks: all dimensions of these four peaks are set to -3, -2, 2 and 3, respectively.
		
		\item Range for each dimension: [-5, 5].
	\end{itemize}
	
	\begin{figure}[h]
		\centering
		\includegraphics[width=0.65\columnwidth]{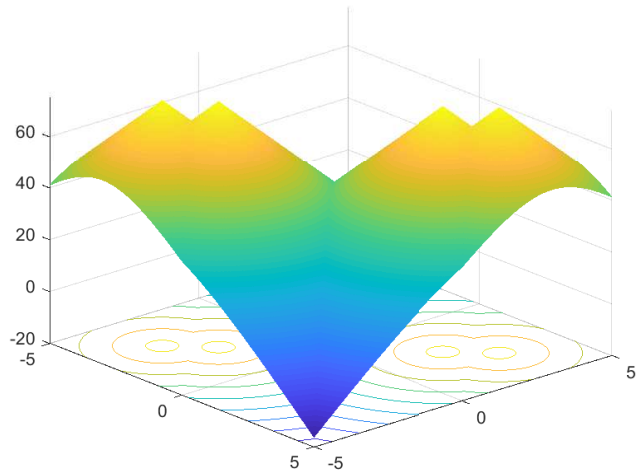}
		\caption{Fitness landscape of F2 in the 2-dimensional problem}	
	\end{figure}

	\subsubsection{DF Function $F_3$}
	
	\begin{itemize}
		\item Property: simulating F7 in CEC 2013 competition on niching methods for multimodal function optimization \cite{li2013benchmark}.
		
		\item Number of global optima: 4.
		
		\item Number of local optima: 0.
		
		\item Widths of peaks: 5.
		
		\item Heights of peaks: 75.
		
		\item Positions of peaks: all dimensions of these four peaks are set to -2.5, -1.5, 0.5 and 4.5, respectively.
		
		\item Range for each dimension: [-5, 5].
	\end{itemize}
	
	\begin{figure}[h]
		\centering
		\includegraphics[width=0.65\columnwidth]{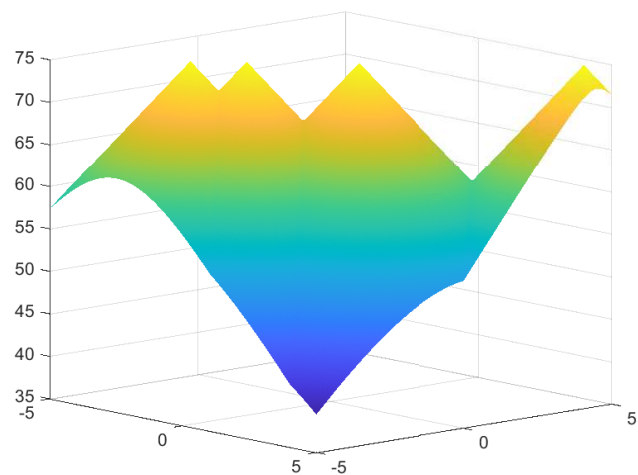}
		\caption{Fitness landscape of F3 in the 2-dimensional problem}	
	\end{figure}
	
	\subsubsection{DF Function $F_4$}
	
	\begin{itemize}
		\item Property: simulating F8 in CEC 2013 competition on niching methods for multimodal function optimization \cite{li2013benchmark}.
		
		\item Number of global optima: 4.
		
		\item Number of local optima: 0.
		
		\item Widths of peaks: 5.
		
		\item Heights of peaks: 75.
		
		\item Positions of peaks: all dimensions of these four peaks are set to -3, -1, 1 and 3, respectively.
		
		\item Range for each dimension: [-5, 5].
	\end{itemize}
	
	\begin{figure}[h]
		\centering
		\includegraphics[width=0.65\columnwidth]{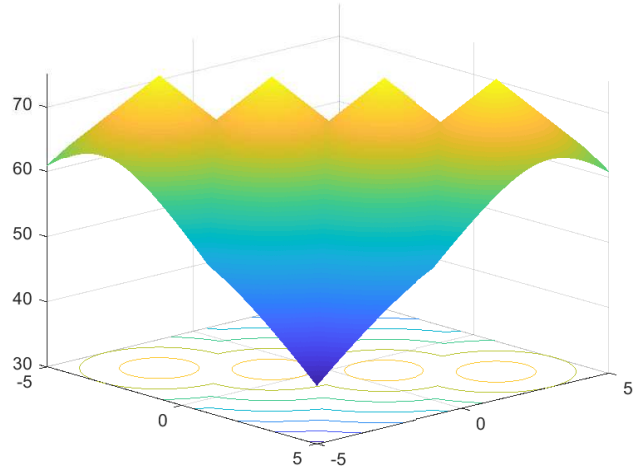}
		\caption{Fitness landscape of F4 in the 2-dimensional problem}	
	\end{figure}
	
	\subsection{Complex Multimodal Functions}
	In the benchmark, the complex multimodal functions are made up of the composition functions in CEC 2013 competition on niching methods for multimodal function optimization \cite{li2013benchmark}. The fitness calculation is defined as follows. 
	
	\begin{equation}
		\label{eqt:comp}
		\begin{split}
			\begin{aligned}
				f&(\boldsymbol{x}) = -\sum_{i=1}^{n} \omega_i (\hat{f}_i((\boldsymbol{x} - \boldsymbol{o}_i)/\lambda_i \cdot M_i))
			\end{aligned}
		\end{split}
	\end{equation}
	
	The composition functions are composed of several basic functions and each optimum of them is the position with the best fitness value. Here, parameter $\omega_i$ is the weight of the $i$-th basic function $f_i$, $\boldsymbol{o}_i$ is the position of global peak in the current basic function, and $\hat{f}_i$ means the normalized function of $f_i$. Parameter $\lambda_i$ controls the scale of the basic functions. When $\lambda_i$ is greater than 1, the landscape is stretched. While it is smaller than 1, the landscape is compressed. $M_i$ is a rotation matrix, which changes the positions.
	
	
	
	\subsubsection{Composition Function $F_5$}
	
	\begin{itemize}
		\item Property: deriving from F9 in CEC 2013 competition on niching methods for multimodal function optimization \cite{li2013benchmark}.
		
		\item Number of global optima: 6.
		
		\item $\lambda = [1, 1, 8, 8, 1/5, 1/5]$. 
		
		\item $\sigma = [1, 1, 1, 1, 1, 1]$.
		
		\item Best fitness value: 0.
		
		\item Range for each dimension: [-5, 5].
	\end{itemize}
	
	\begin{figure}[!h]
		\centering
		\includegraphics[width=0.65\columnwidth]{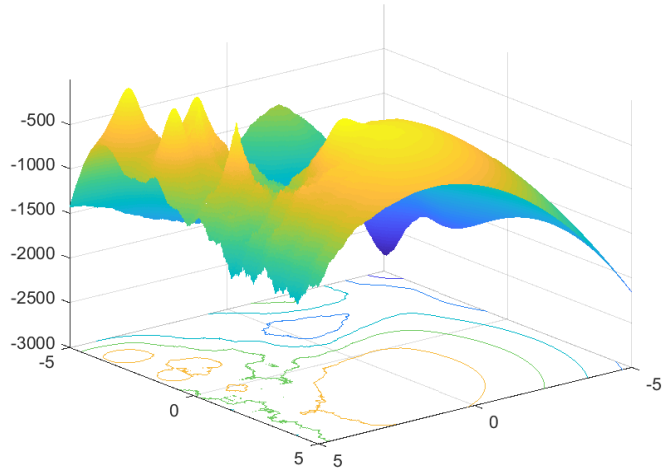}
		\caption{Fitness landscape of F5 in the 2-dimensional problem}	
	\end{figure}
	
	\subsubsection{Composition Function $F_6$}
	
	\begin{itemize}
		\item Property: deriving from F10 in CEC 2013 competition on niching methods for multimodal function optimization \cite{li2013benchmark}.
		
		\item Number of global optima: 8.
		
		\item $\lambda = [1, 1, 10, 10, 1/10, 1/10, 1/7, 1/7]$ .
		
		\item $\sigma = [1, 1, 1, 1, 1, 1, 1, 1]$.
		
		\item Best fitness value: 0.
		
		\item Range for each dimension: [-5, 5].
	\end{itemize}
	
	\begin{figure}[!h]
		\centering
		\includegraphics[width=0.65\columnwidth]{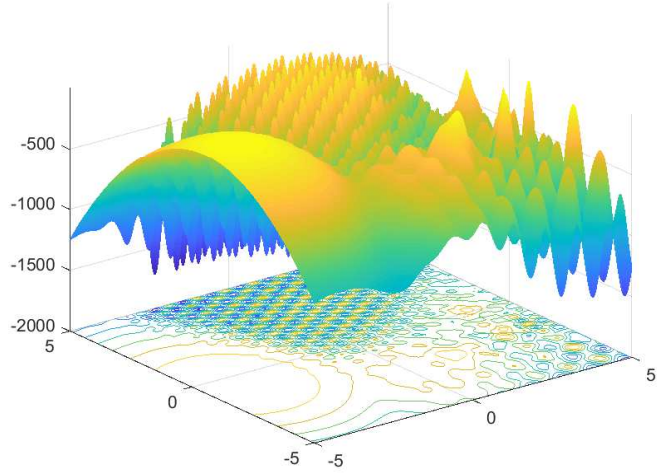}
		\caption{Fitness landscape of F6 in the 2-dimensional problem}	
	\end{figure}
	
	\subsubsection{Composition Function $F_7$}
	
	\begin{itemize}
		\item Property: deriving from F11 in CEC 2013 competition on niching methods for multimodal function optimization \cite{li2013benchmark}.
		
		\item Number of global optima: 6.
		
		\item $\lambda = [1/4, 1/10, 2, 1, 2, 5]$. 
		
		\item $\sigma = [1, 1, 2, 2, 2, 2]$.
		
		\item Best fitness value: 0.
		
		\item Range for each dimension: [-5, 5].
	\end{itemize}
	
	\begin{figure}[!h]
		\centering
		\includegraphics[width=0.65\columnwidth]{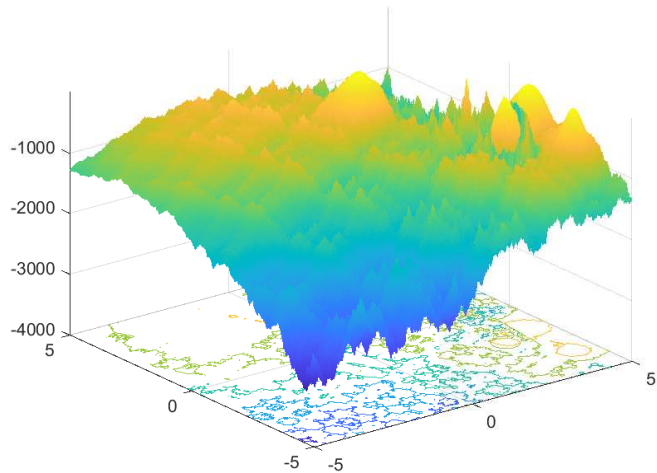}
		\caption{Fitness landscape of F7 in the 2-dimensional problem}	
	\end{figure}
	
	\subsubsection{Composition Function $F_8$}
	
	\begin{itemize}
		\item Property: deriving from F12 in CEC 2013 competition on niching methods for multimodal function optimization \cite{li2013benchmark}.
		
		\item Number of global optima: 8.
		
		\item $\lambda = [4, 1, 4, 1, 1/10, 1/5, 1/10, 1/40]$.
		
		\item $\sigma = [1, 1, 1, 1, 1, 2, 2, 2]$.
		
		\item Best fitness value: 0.
		
		\item Range for each dimension: [-5, 5].
	\end{itemize}
	
	\begin{figure}[!h]
		\centering
		\includegraphics[width=0.65\columnwidth]{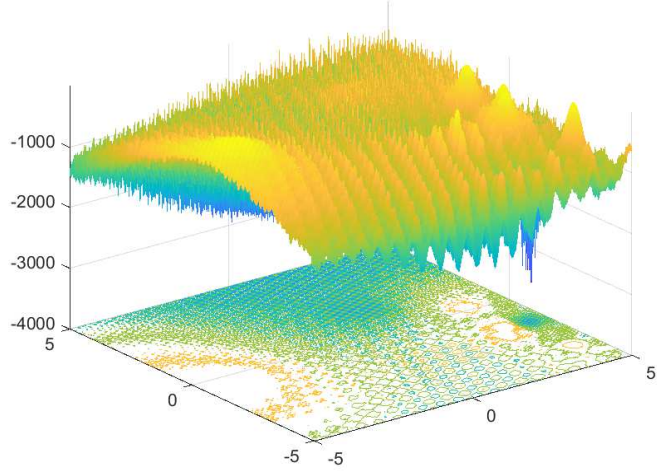}
		\caption{Fitness landscape of F8 in the 2-dimensional problem}	
	\end{figure}

	\section{Dynamic Change Modes}
	\label{sec:dynamic}
	
	In this section, several dynamic change modes are discussed to reveal how the fitness landscape is changed when environmental change occurs. Here, 6 basic change modes about the key parameters proposed in GDBG \cite{li2008benchmark} are introduced in our proposed DMMOPs. Besides, two additional change modes about the number of global optima are given to test the performances of the algorithms.
	
	The changed parameters of different basic multimodal functions are different. Specifically, in the basic multimodal functions from $F_1$ to $F_4$, the heights $H_i$ of the local optima, the widths $W_i$ and the positions $\boldsymbol{X}_i$ of all optima are changed by the corresponding change modes. For the functions from $F_5$ to $F_8$, the parameters related to the positions of optimal solutions $\boldsymbol{o}_i$ and the rotation matrix $M_i$ are changed.
	
	
	Formulas (\ref{eqt:small}) $\sim$ (\ref{eqt:random_peaks}) defined in the following subsections illustrate the change modes. If the parameter is a numerical value, it is changed according to the following formulas. As for the matrix parameter, such as the position of optima $\boldsymbol{X}_i$, $\boldsymbol{o}_i$ and the rotation matrix $M_i$, the rotation angle $\theta$ of the corresponding matrix is changed. Next, the dimension numbers are randomly separated into several groups of which each contains two numbers. It is noted that if the dimension $D$ is an odd number, only $(D-1)$ numbers are randomly selected. Then, the rotation matrix $R$ about the parameters is constructed by the randomly generated dimension numbers and $\theta$. Finally, the parameter in the next environment is changed by $R$.
	
	
	\subsection{Basic Change Modes}
	
	For a given parameter $E(t)$ at the $t$-th environment, the following 6 strategies are adopted to change the environment, which is derived from \cite{li2008benchmark}. The parameters are illustrated in the following part.
	
	\begin{enumerate}
		\item \textit{Change Mode 1} (C1): Small step changes. In this mode, the parameter is slightly changed when the environment changes, which is shown as follows.
		\begin{equation}
			\label{eqt:small}
			E(t+1) = E(t) + \alpha \cdot E_{r} \cdot r \cdot E_{s}
		\end{equation}	
		\item \textit{Change Mode 2} (C2): Large step changes. In this mode, the change step of the parameter is much larger than C1, which is shown as follows.
		\begin{equation}
			\label{eqt:large}
			\begin{split}
				E(t+1) &= E(t) + E_{r} \cdot (\alpha \cdot sign(r)+(\alpha_{max}-\alpha)\cdot r)\cdot E_{s}, \\
				sign(r) &= \begin{cases}
					1, & r > 0; \\
					0 & r = 0; \\
					-1 & r < 0; 
					\end{cases}
			\end{split}
		\end{equation}
		\item \textit{Change Mode 3} (C3): Random changes. In this mode, when the environment changes, the parameter is added to a standard Gaussian distribution value $N(0, 1)$ with a step $E_{s}$.
		\begin{equation}
			\label{eqt:random}
			E(t+1) = E(t) +  E_{s} \cdot N(0, 1)
		\end{equation}
		\item \textit{Change Mode 4} (C4): Chaotic changes. The parameter in this mode is changed chaotically, which is shown in Formula (\ref{eqt:chaotic}).
		\begin{equation}
			\label{eqt:chaotic}
			E(t+1) = E_{min} + A \cdot (E(t)-E_{min}) \cdot (1-(E(t)-E_{min})/E_{s})
		\end{equation}
		\item \textit{Change Mode 5} (C5): Recurrent changes. In this mode, the parameters in all the environments are changed periodically. In other words, the environmental data of the problem is generated in a periodic manner. Formula (\ref{eqt:recurrent}) shows the changes.
		\begin{equation}
			\label{eqt:recurrent}
			E(t+1) = E_{min} + E_r \cdot (sin(\frac{2\pi}{P}t+\varphi)+1)/2
		\end{equation}
		\item \textit{Change Mode 6} (C6): Recurrent with noisy changes. In this mode, the parameters are changed periodically but with some noisy.  There still exist some small difference between the current environment and its most similar one. It means that the current environment would be similar to a certain historical one after several changes.
		\begin{equation}
			\label{eqt:noise}
			\begin{split}
				\begin{aligned}
					E(t+1) =& E_{min} + E_{r} \cdot (sin(\frac{2\pi}{P}t+\varphi)+1)/2 \\
					&+ n_{s} \cdot N(0,1)
				\end{aligned}
			\end{split}
		\end{equation}
	\end{enumerate}
	
	The parameters in Formulas (\ref{eqt:small}) $\sim$ (\ref{eqt:noise}) are set as \cite{li2008benchmark}. The parameters $\alpha$, $\alpha_{max}$, $A$, $P$, and $n_s$ are constants and set to 0.04, 0.01, 3.67, 12 and 0.8, respectively. Parameter $\phi$ is the initial phase and randomly set to a pre-defined fixed value. Parameter $r$ is randomly picked from -1 to 1. Parameter $E_r$ represents the range of the parameter $E(t)$, and it is set in the range from $E_{min}$ to $E_{max}$ which are the lower and upper bounds of $E(t)$, respectively. Parameter $E_s$ means the change severity of the corresponding data $E(t)$. Specifically, the heights and widths of the peaks and rotation angels are directly changed by these formulas. The height changes between 30 and 70 (the severity is 7.0), while the width changes between 1 and 12 (the severity is 1.0). As for the rotation angle, the changing severity is 1 in all the change strategies. The range of the rotation angle is $[0, \pi/6]$ in C5 and C6, and $[-\pi, \pi]$ in other strategies.
	
	
	\subsection{Additional Change Modes}
	Two additional change modes about the number of global optima are introduced into our DMMOPs. The first one is that the number of global optima is changed linearly, while the other is that the number is randomly generated from a specific range.
	
	\begin{enumerate}

		\item \textit{Change Mode 7} (C7): The number of the global optima is changed linearly when the environment changes. In the C7 change mode, the number of global optima of all benchmarks is varied from 2 to the maximum number defined in Section \ref{sec:multi}. The change is shown in Formula (\ref{eqt:periodical_peaks}), where the parameter $dir$ represents that the number of the global peaks in the next environment $g(t+1)$ increases (i.e., $dir = 2$) or decreases (i.e., $dir = 1$) linearly. When the number reaches the maximum number, $dir$ is set to 1 to reduce the number of global optima. While the number goes to 2, $dir$ is fixed to 2 to increase the number when the environment changes. Other parameters such as the key parameters are modified as the \textit{Change Mode 1} (C1).
		\begin{equation}
			\label{eqt:periodical_peaks}
			g(t+1) = g(t) + (-1)^{dir}
		\end{equation}
		
		\item \textit{Change Mode 8} (C8): The number of the global optima changes randomly. Specifically, when the environment change occurs, the number of the global optima are randomly picked in the range from 2 to the initial number of the corresponding functions. Formula (\ref{eqt:random_peaks}) shows the C8 change mode, where $randi(2, g_{max})$ represents the random function to generate the integer between 2 and the given maximum number $g_{max}$. Similar to C7, the rest key parameters are changed by C1.

		\begin{equation}
			\label{eqt:random_peaks}
			g(t+1) = randi(2,\ g_{max})
		\end{equation}
	\end{enumerate}
	
	\subsection{Other Details}
	When the optima move during the environmental changes, the distance between two optima may be very close. To prevent the optima from running into the same position or having a very small distance, the minimum distance $dpeaks$ between these optima is defined. That is, the parameter $dpeaks$ means that the distance between any two optima should be greater than $dpeaks$. If a peak moves to a position that does not satisfy the condition, the peak would move multiple times until the minimum distance is satisfied. The movement of peaks in the rest times is shown in Formula (\ref{eqt:dpeaks}).
	
	\begin{equation}
		\label{eqt:dpeaks}
		\begin{split}
			\boldsymbol{X}_i &= \boldsymbol{X}_i + \boldsymbol{v}, \\
			\boldsymbol{v} &= \frac{\boldsymbol{r}}{\|\boldsymbol{r}\|} * dpeaks,
		\end{split}
	\end{equation}
	where $\boldsymbol{r}$ represents a random direction used in the movement and the length of the movement $\boldsymbol{v}$ is set to $dpeaks$. In our benchmark, $dpeaks$ is set to 0.1.
	
	Besides, the proposed DMMOP has 60 environments and each environment contains $5000*D$ fitness evaluations. 
	
	\section{Experimental Criteria}
	\label{sec:ex}
	
	\subsection{General Settings}
	In this competition, 24 benchmark problems should be tested, which are shown in Table \ref{tab:bench}. The problems are divided into three groups. The problems in Group 1 have different basic multimodal functions, but changes in the same mode when the environment changes. The problems in Group 2 have the same basic multimodal functions, but the change modes are different. The third group is used to test the algorithm performance on the relatively high dimensional problems.
	
	\begin{table}[!h]
		\centering
		\caption{Details of the benchmark problems}
		\begin{tabular}{ccccc}
			\toprule
			\multicolumn{1}{l}{Group} & Index & Multimodal function & Dynamic Mode & Dimension \\
			\midrule
			\multirow{8}[0]{*}{G1} & P1   & F1   & C1   & 5 \\
			& P2   & F2   & C1   & 5 \\
			& P3   & F3   & C1   & 5 \\
			& P4   & F4   & C1   & 5 \\
			& P5   & F5   & C1   & 5 \\
			& P6   & F6   & C1   & 5 \\
			& P7   & F7   & C1   & 5 \\
			& P8   & F8   & C1   & 5 \\
			\midrule
			\multirow{8}[0]{*}{G2} & P9   & F8   & C1   & 5 \\
			& P10  & F8   & C2   & 5 \\
			& P11  & F8   & C3   & 5 \\
			& P12  & F8   & C4   & 5 \\
			& P13  & F8   & C5   & 5 \\
			& P14  & F8   & C6   & 5 \\
			& P15  & F8   & C7   & 5 \\
			& P16  & F8   & C8   & 5 \\
			\midrule
			\multirow{8}[0]{*}{G3} & P17  & F1   & C1   & 10 \\
			& P18  & F2   & C1   & 10 \\
			& P19  & F3   & C1   & 10 \\
			& P20  & F4   & C1   & 10 \\
			& P21  & F5   & C1   & 10 \\
			& P22  & F6   & C1   & 10 \\
			& P23  & F7   & C1   & 10 \\
			& P24  & F8   & C1   & 10 \\
			\bottomrule
		\end{tabular}%
		\label{tab:bench}%
	\end{table}%
	
	The rest settings for DMMOPs are listed as follows.
	
	\begin{enumerate}
		\item \textbf{Runs}: 30. The random seeds of all runs should be fixed from 1 to 30.
		\item \textbf{Frequency}: each environment contains $5000*D$ fitness evaluations.
		\item \textbf{Number of Environments}: 60.
		\item \textbf{Maximum Fitness evaluations}: the sum of fitness evaluations in all environments is set to $5000*D*60$.
		\item \textbf{Environmental change condition}: there is no fitness evaluation for the current environment.
		\item \textbf{Termination condition}: all the fitness evaluations are consumed.
	\end{enumerate}
	
	\subsection{Performance Metric}
	This report adopts the peak ratio (PR) in \cite{lin2021popdmmo} as the indicator to evaluate the performance of different algorithms. The calculation is shown as follows.
	
	\begin{equation}
		\label{eqt:pr}
		PR=\frac{\sum_{i=1}^{Run} \sum_{j=1}^{Env} NPF_{ij}} {\sum_{i=1}^{Run} \sum_{j=1}^{Env} Peaks_{ij}}
	\end{equation}
	
	In Formula (\ref{eqt:pr}), the parameter $Run$ means the total number of the independent algorithm runs, which equals 30. Parameter $Env$ means the number of the environment in a run, which is set to 60. Parameters $NPF_{ij}$ represents the number of optimal solutions in the final population of the $j$-th environment and $i$-th run, while $Peaks_{ij}$ is the total number of global peaks in the corresponding environment and run.
	
	
	%
	%
	
	When calculating the performance metric, it is a core issue to figure out how many global peaks are found by the algorithm, i.e., $NPF_{ij}$ in Formula (\ref{eqt:pr}). Considering that the global optima in each environment of the problem are known, we check whether a solution is a global optimal solution in terms of both fitness values and distances. Algorithm \ref{alg:peaks} shows the detail process. That is, each solution in the final population is iterated. Then, the nearest global optima $\boldsymbol{o}'$ is obtained by the Euclidean distance. Next, the fitness gap and distance gap between the current individual and the nearest global optima $\boldsymbol{o}'$ are calculated. If the distance gap is less than $\epsilon_d$ and the fitness gap is less than $\epsilon_f$, the global optima $\boldsymbol{o}'$ is considered to be found by the algorithm. If $\boldsymbol{o}'$ is not in the optima set $S$,  $\boldsymbol{o}'$ is inserted into $S$. Finally, the size of $S$ is considered as the number of global optima found by the algorithm. Here, the parameter $\epsilon_d$ is set to 0.05, and the parameter $\epsilon_f$ is set to three levels, i.e., 1e-3, 1e-4, and 1e-5, respectively.
	
	
	\begin{algorithm}[h]
		\caption{Obtain the value $NPF$}
		\label{alg:peaks}
		\begin{flushleft}
			\textbf{Input:} $P$ (resultant population), $N$ (resultant population size), $\epsilon_f$ (accuracy), $\epsilon_d$ (distance gap), $F$ (fitness function) \\
			\textbf{Output:} $NPF$ (number of global peaks found by the algorithm)
		\end{flushleft}
		\begin{algorithmic}[1]
			\State Set $S = \emptyset$;
			\For {$i = 1$ \textbf{to} $N$}
			\State Set $\boldsymbol{o}'$ as the global peak nearest to the individual $\boldsymbol{x}_i$;
			\If {$|F(\boldsymbol{x}_i) - F(\boldsymbol{o}')| < \epsilon_f$ \textbf{and} $\|\boldsymbol{x}_i - \boldsymbol{o}'\|< \epsilon_d$}
			\State Consider that the global peak $\boldsymbol{o}'$ is found;
			\If {$\boldsymbol{o}'$ \textbf{not in} $S$}
			\State Add $\boldsymbol{o}'$ into $S$;
			\EndIf
			\EndIf
			\EndFor
			\State Set $NPF$ as the size of $S$;
		\end{algorithmic}
	\end{algorithm}
	
	\subsection{Results Format}
	Participants must submit their $NPF$ results of all benchmark problems with the given table shown as follows. Here, the results with 1e-3, 1e-4 and 1e-5 of $\epsilon_f$ are recorded. In detail, the column named $PR$ records the performance of the optimization algorithm in the corresponding level of accuracy. The columns named Best and Worst represent the found peak ratio in the best and worst runs, respectively. That is, the best and worst ratio results between the peaks found by the algorithm and all the peaks are recorded.
	
	\begin{table}[htbp]
		\centering
		\caption{Results record for solving DMMOPs}
		\begin{tabular}{|c|c|ccc|ccc|ccc|}
			\hline
			\multirow{2}[0]{*}{Group} & \multirow{2}[0]{*}{Index} & \multicolumn{3}{c|}{$\epsilon_f=1e-3$} & \multicolumn{3}{c|}{$\epsilon_f=1e-4$} & \multicolumn{3}{c|}{$\epsilon_f=1e-5$} \\
			&      & PR & Best & Worst & PR & Best & Worst & PR & Best & Worst \\
			\hline
			\multirow{8}[0]{*}{G1} & P1   &      &      &      &      &      &      &      &      &  \\
			& P2   &      &      &      &      &      &      &      &      &  \\
			& P3   &      &      &      &      &      &      &      &      &  \\
			& P4   &      &      &      &      &      &      &      &      &  \\
			& P5   &      &      &      &      &      &      &      &      &  \\
			& P6   &      &      &      &      &      &      &      &      &  \\
			& P7   &      &      &      &      &      &      &      &      &  \\
			& P8   &      &      &      &      &      &      &      &      &  \\
			\hline
			\multirow{8}[0]{*}{G2} & P9   &      &      &      &      &      &      &      &      &  \\
			& P10  &      &      &      &      &      &      &      &      &  \\
			& P11  &      &      &      &      &      &      &      &      &  \\
			& P12  &      &      &      &      &      &      &      &      &  \\
			& P13  &      &      &      &      &      &      &      &      &  \\
			& P14  &      &      &      &      &      &      &      &      &  \\
			& P15  &      &      &      &      &      &      &      &      &  \\
			& P16  &      &      &      &      &      &      &      &      &  \\
			\hline
			\multirow{8}[0]{*}{G3} & P17  &      &      &      &      &      &      &      &      &  \\
			& P18  &      &      &      &      &      &      &      &      &  \\
			& P19  &      &      &      &      &      &      &      &      &  \\
			& P20  &      &      &      &      &      &      &      &      &  \\
			& P21  &      &      &      &      &      &      &      &      &  \\
			& P22  &      &      &      &      &      &      &      &      &  \\
			& P23  &      &      &      &      &      &      &      &      &  \\
			& P24  &      &      &      &      &      &      &      &      &  \\
			\hline
		\end{tabular}%
		\label{tab:addlabel}%
	\end{table}%

	
	
	\bibliography{main}{}
	\bibliographystyle{IEEEtran}
	
\end{document}